# Word Embeddings for Banking Industry

Avnish Patel
avipatel68@gmail.com

March 2023

## Abstract

Applications of Natural Language Processing (NLP) are plentiful, from sentiment analysis to text classification. Practitioners rely on static word embeddings (e.g. Word2Vec or GloVe) or static word representation from contextual models (e.g. BERT or ELMo) to perform many of these NLP tasks. These widely available word embeddings are built from large amount of text, so they are likely to have captured most of the vocabulary in different context. However, how well would they capture domain-specific semantics and word relatedness? This paper explores this idea by creating a bank-specific word embeddings and evaluates them against other sources of word embeddings such as GloVe and BERT. Not surprising that embeddings built from bank-specific corpora does a better job of capturing the bank-specific semantics and word relatedness. This finding suggests that bank-specific word embeddings could be a good stand-alone source or a complement to other widely available embeddings when performing NLP tasks specific to the banking industry.

## 1 Introduction

Word embeddings have been integral in performing NLP tasks for many years. As Bommasani et al. (2020)[1] pointed out that there has been a tremendous advancement in embedding quality by incorporating syntax, morphology, subwords, and context. Central to all of this is the large number of corpora leveraged to develop these embeddings. As such, these embeddings are capturing information about a word and its relationship to other words in a generalized manner.

The central hypothesis of this paper is that word embeddings built from a bank-specific corpora would better capture the domain specific semantics than those built from a broader set of corpora (e.g. Wikipedia, Twitter). For example, words 'credit' and 'line' have a high degree of relatedness in banking and not as much in everyday usage of these two words.

In order to construct bank-specific embeddings, a sizeable and credible corpus specific to banking is necessary. It requires deriving word-word cooccurrence from the corpora. The dimensionality of this cooccurrence matrix needs to be reduced to form the word embeddings that capture the relationship between words. Several techniques need to be explored to identify a set of embeddings that best capture the word relatedness. Pretrained contextual word representation should be considered as well, which requires converting them to static embeddings for the corpora's vocabulary.

As NLP gains more traction and adoption, banking industry can benefit from deploying NLP tasks such as sentiment analysis and information retrieval. These tasks can yield operational effectiveness and efficiency. Bank specific embeddings can be a source that can help improve the performance of NLP tasks in banking.

This paper demonstrates a way to build bank-specific word embeddings using various dimensionality reduction techniques. In order to gauge their effectiveness, static embeddings are obtained for the corpora's vocabulary from several pretrained contextual models for comparison. All these different sets of embeddings are evaluated against a dataset of word-word pairs that were scored by humans for relatedness.

The central finding is that the word embeddings built from bank-specific corpora better captures the word relatedness of the word-word pairs in the test dataset. Of the several techniques that were applied, only a couple outperformed the base model. Further, no set of embeddings that were derived for contextual models outperformed the base model. The findings support the hypothesis that domain specific word embeddings derived from domain related corpora better captures the semantics unique to that domain.

---

[1] Link to: Bommasani et al. (2020)

## 2 Related work

### 2.1 Vector space models of semantics

Most generically, word embeddings are fixed length vector representation for words. Word embeddings can be broken down into two types, prediction-based models and count-based models (Pennington et al. (2014)). This paper explores both methods.

Approaches to word embeddings can be classified as extrinsic evaluation or intrinsic evaluation (Schnabel et al., 2015). In this paper, intrinsic evaluation is the focus so we will evaluate word embeddings against human judgement.

Construction of word embeddings using count-based approach can be broken down into five steps (Turney et. al., 2010):

1. Building the frequency matrix
2. Weighting the elements
3. Smoothing the matrix
4. Comparing the vectors
5. Machine learning

This is the general approach this paper takes as well in constructing word embeddings from banking specific corpora. In this paper, weighting the elements for count-based models uses positive pointwise mutual information (PPMI) method. PPMI, illustrated below, will give a high value of $p_{i,j}$ when there is an interesting semantic relation between the word and the context.

$$p_{ij} = \frac{f_{ij}}{\sum_{i=1}^{n_r} \sum_{j=1}^{n_c} f_{ij}}$$

$$p_{i*} = \frac{\sum_{j=1}^{n_c} f_{ij}}{\sum_{i=1}^{n_r} \sum_{j=1}^{n_c} f_{ij}}$$

$$p_{*j} = \frac{\sum_{i=1}^{n_r} f_{ij}}{\sum_{i=1}^{n_r} \sum_{j=1}^{n_c} f_{ij}}$$

$$\text{pmi}_{ij} = \log\left(\frac{p_{ij}}{p_{i*} p_{*j}}\right)$$

$$x_{ij} = \begin{cases} \text{pmi}_{ij} & \text{if pmi}_{ij} > 0 \\ 0 & \text{otherwise} \end{cases}$$

### 2.2 Static embeddings

This paper also evaluates word embeddings derived from Word2Vec (Mikolov et al., 2013) and GloVe (Pennington et al., 2014). Both of them leveraged large amount of text to build word embeddings. GloVe is a log-bilinear regression model that combines global matrix factorization and local context window. Novel approach with GloVe is that training was done only on the non-zero elements to create a co-occurrence matrix. Word2Vec package includes two log-linear model architectures: continuous Skip-gram model and continuous Bag-of-Words (CBOW) model. In this paper, embeddings from CBOW model will be evaluated as this approach predicts a word given the context around it as illustrated below.

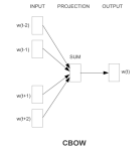

Most recently, contextualized representations (e.g., BERT, ELMo) are widely used for downstream NLP applications. So, this paper will explore the static embeddings derived from BERT using the methods Bommasani et al., (2020) developed. Specifically, this paper uses the decontextualized approach for specifying contexts and combining the associated representations.

## 3 Data

The bank-specific corpora were constructed from the complaints database available on Consumer Financial Protection Bureau (CFPB) website [2]. This is the formal portal for consumers to log complaints against a financial services and related companies. All complaints with a narrative (i.e., written complaint) from Feb 2011 to Feb 2023 were considered. This resulted in total of 1.2MM unique complaints. Table-1 below highlights some key descriptive statistics.

| Number of Comments | 1,213,410 |
|---|---|
| Total Words[1] | 727,069,416 |
| Vocabulary[1] | 173,034 |
| Number of comments with <200 words | 1,122,285 |
| 1 Excludes common stop words | |

Table 1: Descriptive Statistics

---

[2] CFPB website

For this research, a 5% random sample from the 1.2MM comments was taken. This resulted in 4.9MM total words with a vocabulary of 38K. This subset is used to create word embeddings using various models which will be highlighted in the next section.

Next, a co-occurrence matrix is generated from the 38K unique words. This is generated by looping through each comment and generating a context window of 10 (d=10) for each of the words in the sentence. Each context word is scaled n/d where n is the relative position from the focus word; thus, the nearest two context words are assigned a value of 1 and the two farthest on each side are assigned 0.1 as illustrated below. This results in a 38K x 38K matrix. Then, every value in the matrix is weighted using the PPMI method which was highlighted earlier.

|  | Context word | ...... | Context word | Context word | Focus word | Context word | Context word | ...... | Context word |
|---|---|---|---|---|---|---|---|---|---|
| d=10 ---> | 10 | ...... | 2 | 1 | 0 | 1 | 2 | ...... | 10 |
| Scaled n/d ---> | 0.1 | ...... | 0.9 | 1 | 1 | 1 | 0.9 | ...... | 0.1 |

Top 5K most common words out of the 38K total words, were selected to randomly generate 1.8K word pairs. These word pairs were scored by three bankers. Scores range between 0 and 1, where 0 is no relatedness and 1 is completely related. This scored dataset is used to evaluate models discussed in the next section.

## 4 Models

For this paper, the objective is to evaluate word embeddings derived from using different methods and models against the human-annotated relatedness score.

Table 2

| Techniques & Models | Dimensions |
|---|---|
| LSA[1] | 100 |
| LSA | 500 |
| LSA+Autoencoder[5] | 300 |
| GloVe[2] | 100 |
| GloVe[2] | 300 |
| Word2Vec[3] | 300 |
| LSTM[4] | 300 |

[1] Baseline model
[2] Model from Dingwall and Potts 2018
[3] https://radimrehurek.com/gensim/models/word2vec.html
[4] Embedding layer of 300 and one lstm hidden layer with 128 neurons
[5] Reduce with LSA (500D), then Autoencoder with 1 hidden layer

First approach to creating word embeddings used several dimensionality reduction methods highlighted in Table 2. LSA with 100-dimension is the baseline model. This paper utilized GloVe implementation form Dingawall and Potts (2018)[3] to create 100-dimension and 300-dimension embeddings. For Word2Vec, the author uses module implementation from Gensim[4]. The author also used a 500-dimension LSA output to feed it through an autoencoder to produce a reduced 300-dimension embeddings. Finally, a LSTM architecture was used to feed the embedding layer through one hidden layer with 128 nodes and output using softmax activation.

Second approach to generating word embeddings for the 38K vocabulary was to use several pre-trained models to look up the static embeddings. Table 3 highlights the author's choices of the models. This is not an exhaustive list of pretrained models, though these are widely used and understood by many of the practitioners.

Table 3

| Model | Dimensions |
|---|---|
| GloVe[1] | 300 |
| Word2Vec[2] | 300 |
| BERT[3] | 768 |
| ADA-002[4] | 1536 |

[1] https://nlp.stanford.edu/projects/glove/
[2] word2vec-google-news-300 from Gensim downloader
[3] bert-base-cased from transformers library, using decontextualized approach
[4] text-embedding-ada-002 from OpenAI

## 5 Experiments

The primary experiment was to understand how the different sets of embeddings would score the test dataset, the human-annotated dataset. Cosine distance between embeddings of each word pair in the dataset is used to predict the relatedness score using the embeddings from each of the mode. The Spearman correlation coefficient between the annotated score and predicted score is used for evaluating each of the models.

The spearman correlation coefficient is used to rank order each of the model's performance against the baseline (LSA 100-dimension). Further, the author evaluates the difference between pretrained models and models built from co-occurrence matrix. Lastly, the author evaluates the impact of the embedding dimensions.

From table 4, the spearman correlation coefficient for the base model is 0.44. There are only two models that show better performance, LSA 500 (0.49) and LSA + Autoencoder (0.51).

---

[3] Link to Dingwall and Potts (2018)

[4] Link to word2vec module from Gensim

Table 4: Evaluation of Models

| Model | Dimensions | Spearman Correlation |
|---|---|---|
| **Dervied Embeddings** | | |
| LSA[1] | 100 | 0.44 |
| LSA | 500 | 0.49 |
| LSA + Autoencoder | 300 | 0.51 |
| GloVe | 100 | 0.32 |
| GloVe | 300 | 0.42 |
| Word2Vec | 300 | 0.33 |
| LSTM | 300 | 0.26 |
| **Embeddings Lookup** | | |
| GloVe | 300 | 0.41 |
| Word2Vec | 300 | 0.30 |
| BERT | 768 | 0.27 |
| ADA-002 | 1536 | 0.34 |

[1] Baseline model/benchmark

GloVe-300, built from weighted co-occurrence matrix, scored nearly as well as the base model (0.42 vs 0.44). Embeddings from pretrained GloVe word vectors dataset also performed close to the base model (0.41 vs 0.44). Word2Vec, whether modeled or looked up, performed very similarly (0.33 and 0.30). As for the pretrained models, both did not perform as well as the base model. Though, ADA-002 performed much better than BERT. This is likely due to using decontextualized approach in retrieving embeddings from BERT and training set differences between ADA and BERT. Finally, looking at the size of embedding dimensions, both LSA and GloVe models performed better when using larger dimension space (500 vs 100 for LSA, 300 vs 100 for GloVe).

Overall, the top three preforming models, including the base model, were built from the weighted co-occurrence matrix derived from the domain specific corpora. This supports the hypothesis that embeddings built from bank specific corpora should do a better job with capturing the banking specific context.

# 6 Analysis

The results from the experiments showed that the word embedding constructed from domain specific corpora does a better job with word relatedness in that domain. To further dig into the results, the author takes the following approaches:

1. Look at top neighbors for a bank specific term across the models (Table 5)
2. Review three related bank-specific words in terms of distance (Table 6)
3. Create clusters from embeddings for each of the model and determine which model captures some of the top bigrams within the same cluster (Table 7)

Table 5: Top 6 neighbors for the word 'payments' identified by each of the models.

| Nearest Neighbor | LSA 300 Dimension | LSA + Autoencoder 300 Dimensions | GloVe 300 Dimensions | BERT 768 Dimensions | Word2Vec 300 Dimensions | OpenAI ADA 1536 Dimensions | LSTM 300 Dimensions |
|---|---|---|---|---|---|---|---|
| 1 | payment | payment | payment | payment | payment | payment | payment |
| 2 | monthly | monthly | monthly | transactions | pmt | ayment | bills |
| 3 | made | made | made | paid | month | paidments | deposits |
| 4 | late | month | late | fees | forbearance | paypay | pmt |
| 5 | month | late | missed | funds | monthly | quadpay | pay |
| 6 | months | loan | making | pay | pmts | paymnts | statements |

## 6.1 Analysis - Approach 1

A very common term in banking is 'payments', so in table 6 the top six neighbors of this term are listed for many of the models. LSA 300D, LSA + Autoencoder 300D, and LSTM models used the weighted co-occurrence matrix to derive the embeddings; while the other four are pretrained and the embeddings were derived by looking up each word in the vocabulary.
The top performing models in terms of Spearman correlation coefficient, LSA 300D and LSA + Autoencoder 300D, share 5 of the 6 neighbors. The pretrained word embeddings tend to be closer to the 'short form' of the term 'payments' (e.g. pmts, pmt, pay, paid). The intuition behind this is that pretrained models tend to use subwords which are then pooled to derive final embeddings. So, 'pay' and 'pmt' would be closer neighbors than other words. ADA's top six neighbors all have the word 'pay', not surprising given the nature of the model architecture. Interestingly, all these seven models' top neighbor is 'payment' (singular form of payments).

## 6.2 Analysis - Approach 2

The author takes three very related terms in banking and graphs them based on their first two dimensions of their embeddings. This should intuitively tell us how each model's embeddings relate these three words. The words {nsf, fees, overdraft} have a linear relationship in banking. Overdraft and nsf (non-sufficient funds) occur when a customer spends more than what's available in her or his account[5]. Both can result in a fee charged by their bank.

Given the relationship amongst these three words, the author uses the embeddings from

---
[5] Overdraft and nsf definition

LSA + Autoencoder, BERT, and ADA to assess how closely these models connect them.

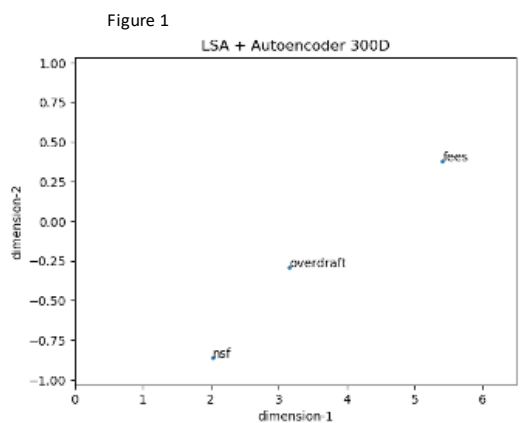

Figure 1

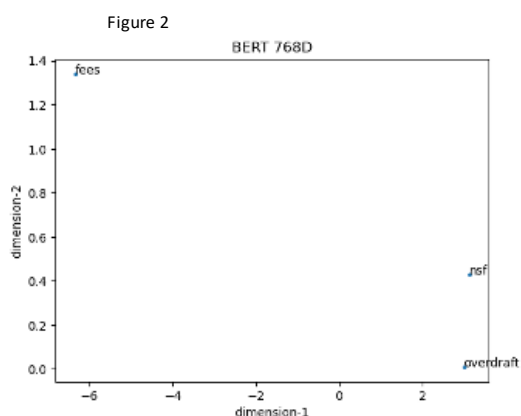

Figure 2

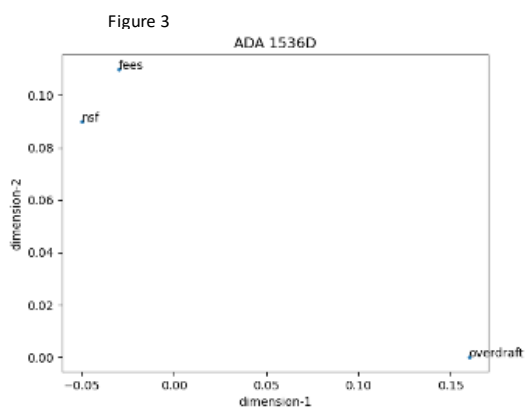

Figure 3

Overall, LSA + Autoencoder pulls these three terms closest to each other; intuitively, this makes sense given that the embeddings were derived from the bank corpora using the weighted co-occurrence matrix. Given the nature of the corpora, there will be more complaints about overdraft fees than about nsf fees because the former tends to have higher incident rate. This is brought out by the LSA + Autoencoder model where the word 'fees' is closer to 'overdraft' than 'nsf'.

BERT shows 'overdraft' and 'fees' much farther away, however, 'nsf' and 'overdraft' are much closer. As for ADA, 'fees' and 'nsf' are closer but both are farther away from 'overdraft'. Intuitively, LSA + Autoencoder keeps all three closer which is most realistic.

### 6.3 Analysis - Approach 3

Table 8: Clustering results

| Word-1 | Word-2 | Model 1 | Model 2 | Model 3 | Model 4 |
|---|---|---|---|---|---|
| home | equity | 1 | 0 | 0 | 0 |
| credit | report | 1 | 1 | 0 | 0 |
| credit | card | 0 | 0 | 0 | 1 |
| identity | theft | 1 | 1 | 0 | 0 |
| fair | credit | 1 | 0 | 0 | 0 |
| reporting | act | 1 | 1 | 0 | 1 |
| account | number | 1 | 1 | 0 | 0 |
| wells | fargo | 1 | 0 | 0 | 0 |
| credit | score | 0 | 1 | 0 | 0 |
| bank | america | 1 | 0 | 0 | 0 |
| inaccurate | information | 1 | 0 | 0 | 0 |
| credit | bureau | 1 | 1 | 0 | 0 |
| collection | agency | 1 | 0 | 1 | 0 |
| third | party | 1 | 1 | 0 | 0 |
| checking | account | 0 | 1 | 1 | 0 |
| please | remove | 1 | 0 | 0 | 0 |
| late | payments | 0 | 0 | 0 | 0 |
| debt | collection | 1 | 0 | 0 | 0 |
| bank | account | 0 | 0 | 0 | 1 |
| late | payment | 0 | 0 | 0 | 0 |
| interest | rate | 1 | 1 | 0 | 1 |
| payment | history | 0 | 1 | 0 | 0 |
| late | fees | 0 | 0 | 0 | 0 |
| home | loan | 1 | 0 | 0 | 0 |
| bank | branch | 1 | 1 | 0 | 0 |
| Accuracy (%) | | 68% | 44% | 8% | 16% |

Table:
Model 1    LSA+Autoencoder
Model 2    BERT
Model 3    ADA
Model 4    GloVe 300d (pretrained word vectors)

When a word pair has a high relatedness score, they should be bound together when the entire vocabulary is clustered into different groups. Thus, if each set of embedding were clustered, the highly related word pairs should end up in the same cluster. With this intuition, the author clustered word embeddings from four models. For each set, a k-means clustering algorithm was applied to create 10 clusters. The author also selected 25 banking related word pairs, with a human-annotated score of 0.8 or higher, from the original 1.8K test dataset. Next, the cluster associated with each word in the word pair is compared to ascertain whether they belong to the same cluster; for example, does word pair W1 & W2 end up in the same cluster? A score of 1 is assigned if both words in the pair end up in the same cluster, otherwise a 0 is assigned.

The four models selected are:

1. LSA + Autoencoder, the best performing model

2. BERT, de facto model
3. ADA from Open AI, another de facto model
4. GloVe 300D, pretrained word vectors built from 6BN tokens with 400K vocabulary

The results in table 8 shows that LSA + Autoencoder outperformed the other three models. This model's embeddings do a good job of capturing the relationship of these highly related word pairs. 68% of the word pairs ended up in the same clusters, next best was BERT with 44% of the word pairs ending up in the same clusters.

Interestingly, three word pairs were not clustered together by any of the model. The common word in these word pair is 'late'; the other words in these pairs were 'payment', 'payments', 'fees'. These are some of the most common word pairs used in customer complaints in the banking industry.

Overall, the embeddings derived from the weighted co-occurrence matrix did the best job of clustering these word pairs into similar clusters.

# 7 Conclusion

This paper hypothesized that word embeddings derived from bank corpora would better capture the semantics associated with banking than pretrained models or word vectors. For example, words 'home' and 'equity' are highly related. In the clustering analysis, the only model to correctly place both words in the same cluster used the co-occurrence matrix developed from bank corpora. Although the findings are not surprising, it does encourage further considerations. The author suggests that following efforts may yield incremental findings and opportunities for improving NLP outcomes in the banking domain. Future work to expand scope:

1. For this research, the author conducted minimal amount of hyperparameter tuning of the models built from the weighted co-occurrence matrix. For example, try different dimensions or different learning rates, etc.
2. Apply other weighting techniques for co-occurrence matrix, such as t-test reweighting.
3. Add additional open-source bank related corpora, such as banking related sources on Wikipedia.
4. Incorporate the embeddings derived directly from the bank corpora with pretrained models to assess impact on NLP tasks like sentiment analysis and text summarization. For example, leverage pretrained contextual models to provide embeddings for out of vocabulary words.
5. Utilize more recent pretrained models in the study, such as GPT3.5.
6. Solicit input from other practitioners on additional ideas that may lead to overall improvement in banking related NLP initiatives.

# 8 Acknowledgements

First, the author is grateful to Professor Potts, Stanford University, for the inspiration and guidance. The author appreciates input, guidance, and motivation provided by Raul Salles de Padua. This paper would not have been possible without Jennifer Brommer and Amy Hartenstine, who were responsible for the human annotated word relatedness scores.

Finally, the author acknowledges that the CFPB's complaints database was the inspiration behind this hypothesis. Hopefully, others will become aware of this wonderful resource for their NLP endeavors.

# 9 Authorship

This paper was written by Avnish 'Avi' Patel based on guidance and instruction from Professor Potts. The source code for creating the co-occurrence matrix and conducting experiments were created by the author based on guidance from several online sources. This paper along with related assets can be found at: https://github.com/mktaop/Word_Embeddings_for_Banking.